\documentclass[11pt,a4paper]{article}
\usepackage[hyperref]{acl2019}
\usepackage{times}
\usepackage{latexsym}
\usepackage{url}

\usepackage{CJKutf8}
\usepackage{booktabs}
\usepackage{makecell}
\usepackage{multirow}
\usepackage{amssymb}
\usepackage{amsmath}
\usepackage{caption}
\usepackage{graphicx, subfig}
\usepackage{color}
\usepackage{hyperref}
\aclfinalcopy % Uncomment this line for the final submission
\title{COS960: A Chinese Word Similarity Dataset of 960 Word Pairs}

\author{Junjie Huang$^{2*\dag}$, 
Fanchao Qi$^{1}\thanks{\ \  Indicates equal contribution}$\hspace{0.2em}, 
Chenghao Yang$^{3}\thanks{\ \ Work done during internship at Tsinghua University}$\hspace{0.2em}, 
Zhiyuan Liu$^{1}$,
Maosong Sun$^{1}$ \hspace{0.5em}\\
 $^{1}$Department of Computer Science and Technology, Tsinghua University \\
 Institute for Artificial Intelligence, Tsinghua University \\
  State Key Lab on Intelligent Technology and Systems, Tsinghua University\\
$^{2}$School of ASEE, Beihang University \quad
$^{3}$Software College, Beihang Unviersity\\
{\tt qfc17@mails.tsinghua.edu.cn, \{hjj1997,alanyang\}@buaa.edu.cn}\\
{\tt \{liuzy,sms\}@tsinghua.edu.cn}
}

\date{}

\begin{document}
\maketitle
\begin{abstract}
 Word similarity computation is a widely recognized task in the field of lexical semantics. Most proposed tasks test on similarity of word pairs of single morpheme, while few works focus on words of two morphemes or more morphemes. In this work, we propose COS960, a benchmark dataset with 960 pairs of Chinese wOrd Similarity, where all the words have two morphemes in three Part of Speech (POS) tags with their human annotated similarity rather than relatedness. We give a detailed description of dataset construction and annotation process, and test on a range of  word embedding models. The dataset of this paper can be obtained from \url{https://github.com/thunlp/COS960}.
\end{abstract}

\section{Introduction}
Word similarity computation is a task to automatically compute similarity score between given word pairs, which is the most popular way to evaluate quality of word embeddings. \citep{Faruqui2016ProblemsWE} The task evaluates the correlation between model computed similarities and human judgement, where the higher correlation is, the more semantic information is captured by the model \citep{Bakarov2018survey}.

There are a large number of diverse dataset constructed to evaluate word similarity, most of which in English. \citet{Rubenstein1965RG65} make an attempt to compute word similarities in order to test the distributional hypothesis \citep{Harris195} and construct the first dataset RG65 including a list of 65 pairs of nouns with their human annotated similarity scores in range of 0-4. After that a series of similarity datasets come out with unique charateristics, including: 

(1) focusing on word relatedness: WordSim-353 \citep{Finkelstein2001ws353},  YP-130 \citep{Yang2006yp130}, MEN \citep{Bruni2012MEN}, MTurk-287 \citep{Radinsky2011mturk287}, MTurk-771 \citep{Halawi2012mturk771}; 

(2) focusing on word true simialrity: SimLex-999 \citep{Hill2015SimLex999ES} , Simverb3500 \citep{Gerz2016SimVerb3500}, Verb-143 \citep{Baker2014verb143}; 

(3) in Chinese: WordSim-297\citep{Jin2012ws297}, WordSim-240 \citep{Wang2011ws240}, polysemous word \citep{Guo2014polyse}, PKU-500 \citep{Wu2016pku500}; 

(4) other highlights: two-word phrasal similarity \citep{Mitchell2010CompositionID}, rare words \citep{Luong2013RW}, words in sentential context \citep{Huang2012ImprovingWR}, cross-lingual word similarity \citep{CamachoCollados2017multilingual}, et.al.

In English, there are a number of datasets focusing on word \textit{true similarity} which has wide applications on dictionary generation \citep{Cimiano2005DictGen}, machine translation \citep{He2008MT, Marton2009ImprovedSMT} and language correction \citep{Li2006LC}. However, such a dataset focusing on word true similarity has been absent in Chinese for a long time. In addition, most of the datasets consist of single-word pairs, few of them consider the similarity of Multiword Expressions (MWEs) which is considered as a "pain in the neck" \citep{Sag2002MultiwordEA} for natural language processing (NLP). 

% \citet{Mitchell2010CompositionID} describe a two-word phrasal similarity dataset which contains 324 pairs of two-word MWEs together with human annotations of pairwise similarity scores from 1 to 7, and the 324 pairs of MWEs are divided into three groups according to their types of combination rule, i.e., \textit{adj-noun}, \textit{noun-noun} and \textit{verb-object}. 

% In fact, MWEs widely appear in different languages, so does Chinese. However, such a benchmark dataset for Chinese MWE similarity computation has been absent for a long time, which becomes a bottleneck in the field of MWEs. To address the issue, we propose COS960, a Chinese wOrd Similarity dataset which contains 960 pairs of words and their human annotated similarity scores. The selected words are all MWEs with two component words and are further grouped according to their Part of Speech (POS) tags. 
% there are 480 pairs of nouns, 240 pairs of verbs and 240 pairs of adjectives separately.

Therefore, we introduce our COS960, a Chinese word similarity dataset of 960 word pairs, where each word is actually a two-word MWE. Each of the word pairs is annotated by 15 native speaker according to its true similarity rather than association. We also report the performance of a variety of word embeddings methods on our COS960 dataset. We hope our COS960 dataset can be helpful in NLP community.

% \section{Related Works}

\section{Dataset Construction}

\subsection{Data Preparation}
\subsubsection*{Word Selection}
 To make sure our word pairs of two morphemes are truly existing Chinese words, we use a famous linguistic knowledge base HowNet as the source of words. We extract the word whose two morphemes and itself all appear in HowNet and form a dataset of such triples in a total number of 51,034.
 
 Then we split the dataset into four parts based on the POS tags of words, which are \textit{noun}, \textit{verb}, \textit{adjective} and \textit{other}. We use their POS tags annotated in HowNet and filter out the words which have more than one POS tags or no POS tag. The final number of each set is $30355,12847, 3603, 4229$ correspondingly. Here we only use the \textit{noun}, \textit{verb} and \textit{adjective} sets.

\subsubsection*{Word Pair Generation}
 We pair the words in the each of the three above-mentioned sets pair by pair. Then we calculate the cosine similarity of each pair based on the word embeddings learned by GloVe \citep{pennington2014glove} in Sogou-T corpus \footnote{Sogou-T is a corpus of web pages containing 2.7 billion words. \url{https://www.sogou.com/labs/resource/t.php}}, and the dimension of word vectors is 200.
 
 We further divide the word sets with three POS tags into five parts respectively according to the similarity range, including $ [1.0, 0.9], [0.9, 0.8], [0.8, 0.7], [0.7, 0.6], [0.6, 0.4]$. Note that we don't take word pairs with cosine similarity lower than 0.4 into account because almost all the them are not really similar to each other. The number of word pairs in each set is shown in Table \ref{tab:data}. Finally, we obtain 480 noun pairs, 240 verb pairs and 240 verb pairs. 

\begin{table}[htbp]
  \centering
    \resizebox{\linewidth}{!}{
    \begin{tabular}{cccc}
    \toprule
          & noun-noun & verb-verb & adjective-adjective \\
    \midrule
    $[1.0, 0.9]$     & 96    & 48    & 48 \\
    $[0.9, 0.8]$     & 96    & 48    & 48 \\
    $[0.8, 0.7]$     & 96    & 48    & 48 \\
    $[0.7, 0.6]$     & 96    & 48    & 48 \\
    $[0.6, 0.4]$     & 96    & 48    & 48 \\
    \midrule
    total & 480   & 240   & 240 \\
    \bottomrule
    \end{tabular}%
    }
    \caption{Number of MWE pairs with different cosine similarities in three sets.}
    \label{tab:data}%
\end{table}%

\subsection{Annotation Details}
The total 960 pairs are randomly shuffled and divided into two parts, each of which contains 480 pairs of data. We recruit 30 native university students, and each of them is asked to annotate 480 pairs of words. 
Annotators are shown the definitions of each word and the categories in TongYiCiCiLin as the reference and are asked to rate a similarity score in a range of 0-4 for each word pair.

Before formal annotation, annotators are asked to read the Annotation Guidebook which presents the differences of \textit{similarity} and \textit{relatedness}.
To improve annotation quality, they are obliged to take an exam before annotating COS960, which consists of at least two word pairs for each POS tag and similarity level (35 in total).

During the process of annotation, they are welcome to discuss and raise questions when they are hesitating, which helps to advance the consistency of different annotation and improve annotation quality.

\subsection{Post-processing}
We calculate the Krippendorff's alpha between each two of the annotators and all their annotation is accepted. Finally, we use the average score of a single pair as the final similarity score and form our COS960. 

\section{Experiment}
In this section, we provide experimental results of several existing word embedding models on our COS960 dataset.

\subsection{Experimental Settings}
% \subsubsection*{Models}
To evaluate our COS960, we choose some typical word embedding models to test including: (1) Skip-Gram \cite{Mikolov2013sgcb}; (2) continuous-bag-of-words (CBOW) \cite{Mikolov2013sgcb}; (3) GloVe \cite{pennington2014glove}; (4) CWE, a character-enhanced word embedding \cite{Chen2015cwe}; (5) fasttext, enriched word vectors with subword information \cite{bojanowski2016fasttext}; (6) cw2vec, a chinese embedding with stroke n-gram information \cite{Cao2018cw2vec}. For hyper-parameters, we set training epochs of every model to 5 and maintain the other default parameters of each model. 

% \subsubsection*{Evaluation Protocol}
For evaluation protocol, we calculate the Pearson correlation coefficient, Spearman’s rank correlation coefficient and the square root of Pearson and Spearman’s rank correlation between cosine similarities of word pairs computed by word embeddings of models and human-annotated scores. 

\subsection{Experimental Results}

\subsubsection*{Overall Results}
 The overall evaluation results on COS960 are shown in Table \ref{tab:960}. From the table, we observe that: 
 
\begin{table}[htbp]
 \resizebox{\linewidth}{!}{
  \centering
    \begin{tabular}{c|ccc}
    \toprule
    \multicolumn{1}{c}{} & Spearman's & Pearson & Square-Mul \\
    \midrule
    Skip-Gram & 76.2  & 71.0   & 73.6 \\
    CBOW  & \textbf{78.2}  & \textbf{72.1}  & \textbf{75.1} \\
    GloVe & 75.0  & 72.0  & 73.5 \\
    CWE   & 72.1  & 65.9  & 69.0 \\
    cw2vec & 75.4  & 68.1  & 71.7 \\
    fasttext & 75.5  & 70.0    & 72.7 \\
    \bottomrule
    \end{tabular}%
}
    \caption{Spearman's rank correlation coefficient ($\rho \times 100$) between similarity scores assigned by compositional models with human ratings on all 960 pairs of words.}
 \label{tab:960}%
\end{table}%

 (1) CBOW achieves the best performance, which is better than the second best model by 2.1\% on average. 
 
 (2) All six methods have considerably high correlation scores with three evaluation protocols. This indicates that the cosine similarity of six evaluated word embeddings still correlates well with word true similarity, which contradicts with \citet{Hill2015SimLex999ES}.
 
 (3) All six methods achieve highest score with the evaluation protocol of Spearman's rank correlation. We attribute it to high annotation consistency that there are often more one word pairs in each similarity level.

% \begin{table}[htbp]
%  \resizebox{\linewidth}{!}{
%   \centering
%     \begin{tabular}{c|ccc|c}
%     \toprule
%     \multicolumn{1}{c}{} & Adj.  & Verb  & Noun  & All \\
%     \midrule
%     Skip-Gram & 80.0  & 83.2  & 74.5  & 76.2 \\
%     CBOW  & 78.5  & 84.8  & 77.0  & 78.2 \\
%     GloVe & 77.7  & 78.5  & 73.7  & 75.0 \\
%     CWE   & 71.6  & 78.1  & 74.2  & 72.1 \\
%     cw2vec & 77.0  & 82.5  & 73.7  & 75.4 \\
%     fasttext & 78.8  & 82.9  & 74.9  & 75.5 \\
%     \bottomrule
%     \end{tabular}%
% }
%   \caption{Spearman's rank correlation }
%  \label{tab:spearman_only}%
% \end{table}%

\subsubsection*{Effect of POS tags}

\begin{table}[htbp]
\resizebox{\linewidth}{!}{
  \centering
    \begin{tabular}{c|ccc}
    \toprule
    \multicolumn{1}{c}{} & Spearman's & Pearson & Square-Mul \\
    \midrule
    Skip-Gram & 74.5  & 66.8  & 70.5 \\
    CBOW  & \textbf{77.0 } & \textbf{69.7}  & \textbf{73.2} \\
    GloVe & 73.7  & 68.6  & 71.1 \\
    CWE   & 74.2  & 64.2  & 69.0 \\
    cw2vec & 73.7  & 64.8  & 69.1 \\
    fasttext & 74.9  & 66.4  & 70.5 \\
    \bottomrule
    \end{tabular}%
}
    \caption{Spearman's rank correlation coefficient ($\rho \times 100$) between similarity scores assigned by compositional models with human ratings on all 480 pairs of nouns.}
\label{tab:noun480}%
\end{table}%

We further present the performance of on COS960 in three POS tags, i.e. nouns in Table \ref{tab:noun480}, verbs in Table \ref{tab:verb240} and adjectives in Table \ref{tab:adj240}.

\begin{table}[htbp]
\resizebox{\linewidth}{!}{
  \centering
    \begin{tabular}{c|ccc}
    \toprule
    \multicolumn{1}{c}{} & Spearman's & Pearson & Square-Mul \\
    \midrule
    Skip-Gram & 83.2  & 81.1  & 82.1 \\
    CBOW  & \textbf{84.8}  & \textbf{80.7}  & \textbf{82.7} \\
    GloVe & 78.5  & 78.1  & 78.3 \\
    CWE   & 78.1  & 76.6  & 77.3 \\
    cw2vec & 82.5  & 78.1  & 80.3 \\
    fasttext & 82.9  & 80.5  & 81.7 \\
    \bottomrule
    \end{tabular}%
}
  \caption{Spearman's rank correlation coefficient ($\rho \times 100$) between similarity scores assigned by compositional models with human ratings on all 240 pairs of verbs.}
\label{tab:verb240}%
\end{table}%

\begin{table}[htbp]
\resizebox{\linewidth}{!}{
  \centering
    \begin{tabular}{c|ccc}
    \toprule
    \multicolumn{1}{c}{} & Spearman's & Pearson & Square-Mul \\
    \midrule
    Skip-Gram & \textbf{80.0}  & \textbf{77.0}  & \textbf{78.5} \\
    CBOW  & 78.5  & 74.4  & 76.4 \\
    GloVe & 77.7  & 76.8  & 77.1 \\
    CWE   & 71.6  & 67.9  & 69.8 \\
    cw2vec & 77.0  & 70.5  & 73.7 \\
    fasttext & 78.8  & 76.1  & 77.5 \\
    \bottomrule
    \end{tabular}%
}
  \caption{Spearman's rank correlation coefficient ($\rho \times 100$) between similarity scores assigned by compositional models with human ratings on all 240 pairs of adjectives.}
  \label{tab:adj240}%
\end{table}%

From Table \ref{tab:noun480}, \ref{tab:verb240} and \ref{tab:adj240}, we find that:

(1) CBOW still performs best in nouns and verbs, which is consistent with overall results,

%%%%%%%%%%%%%%%%%%%%%%%%%%%%%%%%%%%%%
(2) Models have best average performance on verb pairs while perform worst on noun pairs. 
%%%%%%%%%%%%%%%%%%%%%%%%%%%%%%%%%%%%%%

\section{Conclusion}
In this paper we propose COS960, a Chinese word similarity dataset of 960 word pairs, where all selected words are MWEs with two component words. We also describe the process of the  dataset construction in detail and perform evaluation on existing word embedding models. We hope this dataset will contribute to the development of distributional semantics in Chinese.

\bibliography{acl2019}
\bibliographystyle{acl_natbib}
\end{document}